\documentclass{article}

\usepackage[preprint]{neurips_2025}
\usepackage[utf8]{inputenc} % allow utf-8 input
\usepackage[T1]{fontenc}    % use 8-bit T1 fonts
\usepackage{hyperref}       % hyperlinks
\usepackage{url}            % simple URL typesetting
\usepackage{booktabs}       % professional-quality tables
\usepackage{amsfonts}       % blackboard math symbols
\usepackage{nicefrac}       % compact symbols for 1/2, etc.
\usepackage{microtype}      % microtypography
\usepackage{xcolor}         % colors

\usepackage{tabularx}
\usepackage{longtable}
\usepackage{enumitem}
\usepackage{graphicx}
\usepackage{parskip}
\usepackage{array}
\usepackage{comment} % For block commenting
\usepackage{acro}
\usepackage{algorithm}
\usepackage{algpseudocode}
\usepackage{multicol}

\DeclareAcronym{mtb}{
  short         = MTB ,
  long          = Molecular Tumor Board ,
  short-plural  = s ,
  long-plural   = s
}

\DeclareAcronym{hao}{
  short = HAO ,
  long  = Healthcare Agent Orchestrator
}

\DeclareAcronym{llm}{
  short         = LLM ,
  long          = Large Language Model ,
  short-plural  = s ,
  long-plural   = s
}

\DeclareAcronym{ehr}{
    short = EHR ,
    long  = Electronic Health Record
}

\DeclareAcronym{mcp}{
    short = MCP ,
    long  = Model Context Protocol
}

\title{Demo: Healthcare Agent Orchestrator (HAO) for Patient Summarization in Molecular Tumor Boards}
\workshoptitle{GenAI4Health@NeurIPS 2025}

\author{\normalfont
Matthias Blondeel\thanks{Equal contribution.} ~\thanks{We gratefully recognize additional contributors in Appendix~\ref{app:contributors}.} \\ Microsoft Health and Life Sciences\thanks{Correspondence: \texttt{hlsfrontierteam@microsoft.com}} \and
Noel Codella\footnotemark[1] \\ Microsoft Health and Life Sciences \and
Sam Preston\footnotemark[1] \\ Microsoft Health and Life Sciences \and
Hao Qiu\footnotemark[1] \\ Microsoft Health and Life Sciences \and
Leonardo Schettini\footnotemark[1] \\ Microsoft Health and Life Sciences \and
Frank Tuan\footnotemark[1] \\ Microsoft Health and Life Sciences \and
Wen-wai Yim\footnotemark[1] \\ Microsoft Health and Life Sciences \and
Smitha Saligrama \\ Microsoft Health and Life Sciences \and
Mert Öz \\ Microsoft Health and Life Sciences \and
Shrey Jain \\ Microsoft Health and Life Sciences \and
Matthew P. Lungren \\ Microsoft Health and Life Sciences \and
Thomas Osborne \\ Microsoft Health and Life Sciences
}
\date{GenAI for Health Workshop @ NeurIPS 2025}

\begin{document}

\maketitle

\begin{abstract}

% Noel suggested reframing:

\acp{mtb} are multidisciplinary forums where oncology specialists collaboratively assess complex patient cases to determine optimal treatment strategies. A central element of this process is the patient summary, typically compiled by a medical oncologist, radiation oncologist, or surgeon, or their trained medical assistant, who distills heterogeneous medical records into a concise narrative to facilitate discussion. This manual approach is often labor-intensive, subjective, and prone to omissions of critical information.  To address these limitations, we introduce the \emph{\textbf{\ac{hao}}}, a \ac{llm}-driven AI agent that coordinates a multi-agent clinical workflow to generate accurate and comprehensive patient summaries for \acp{mtb}. Evaluating predicted patient summaries against ground truth presents additional challenges due to stylistic variation, ordering, synonym usage, and phrasing differences, which complicate the measurement of both succinctness and completeness. To overcome these evaluation hurdles, we propose \emph{\textbf{TBFact}}, a ``model-as-a-judge'' framework designed to assess the comprehensiveness and succinctness of generated summaries. Using a benchmark dataset derived from de-identified tumor board discussions, we applied TBFact to evaluate our Patient History agent. Results show that the agent captured 94\% of high-importance information (including partial entailments) and achieved a TBFact recall of 0.84 under strict entailment criteria. We further demonstrate that TBFact enables a data-free evaluation framework that institutions can deploy locally without sharing sensitive clinical data. Together, \ac{hao} and TBFact establish a robust foundation for delivering reliable and scalable support to \acp{mtb}.

\end{abstract}

\acreset{mtb}
\acreset{hao}

\section{Introduction}

\acp{mtb} are among healthcare’s most complex collaborative workflows, where radiologists, pathologists, oncologists, geneticists, and other specialists align on patient-specific cancer treatment strategies. The backbone of these meetings is a comprehensive yet concise patient summary that distills heterogeneous records (clinical notes, imaging, pathology, genomics) into a coherent timeline that orients discussion and decision-making. However, manual preparation of these summaries imposes a significant time burden: radiologists and pathologists report mean preparation times of 81.7 and 144.0 minutes respectively~\cite{asco_pilot_study}. Quality is also variable; the study notes that the inclusion of comorbidities and patient perspectives often falls below average standards, and the quality of case history and pathological information remains inconsistent~\cite{asco_pilot_study}.

While general-purpose large language models have demonstrated impressive capabilities across many domains, they face critical limitations in high-stakes healthcare settings. Precision is critical: even minor hallucinations or inconsistencies can compromise patient safety and decision quality. Effective decision-making also requires multi-modal integration—correlating imaging, pathology, genomics, and structured \ac{ehr} data—much of which lies outside the scope of public training corpora. Finally, transparency and traceability are essential: clinicians must understand how conclusions are formed and be able to audit intermediate steps. 

To address these challenges, we introduce the \emph{\textbf{\acl{hao}}}, a modular, \ac{llm}-driven multi-agent system that mirrors the collaborative structure of real tumor boards. Rather than relying on a single monolithic model, \ac{hao} coordinates specialized agents—each focused on a domain such as patient history, radiology, pathology, staging, or clinical guidelines—while maintaining coherence and explainability through an orchestrator. This design enables grounded reasoning across heterogeneous data sources and aligns with the multidisciplinary nature of oncology care, allowing \ac{hao} to expand into other use-cases.

% Evaluating the quality of generated patient summaries introduces its own challenges. Common similarity metrics that emphasize surface overlap are insensitive to clinically meaningful differences: two summaries may use different phrasing, ordering, or synonyms while conveying the same facts—or conversely, appear similar while omitting critical details or introducing unsupported claims. To address this gap, we propose \emph{\textbf{TBFact}}, a claim-level, ``model-as-a-judge'' framework that decomposes summaries into clinical factual claims facts and applies bidirectional entailment to quantify both \emph{completeness} (coverage of reference facts) and \emph{succinctness} (a precision-oriented proxy that penalizes unsupported or extraneous statements), while attributing mismatches to \emph{omissions} or \emph{unsupported claims} for actionable diagnostics. Importantly, TBFact supports institution-controlled deployment: evaluation can run without transmitting raw patient data outside organizational boundaries, and institutions may configure either on-premise models or private/vended endpoints according to their governance requirements.

Evaluating the quality of generated patient summaries introduces its own challenges. Common similarity metrics that emphasize surface overlap are insensitive to clinically meaningful differences: two summaries may use different phrasing, ordering, or synonyms while conveying the same facts—or conversely, appear similar while omitting critical details or introducing unsupported claims. We therefore implement \emph{\textbf{TBFact}}, an evaluation framework that operates at the level of \emph{clinical factual claims}, and assess bidirectional entailment between candidate and reference to quantify both \emph{completeness} (coverage of reference facts) and \emph{succinctness} (a precision-oriented proxy that penalizes unsupported or extraneous statements). 

\section{\acf{hao}}
\label{sec:hao}

% Domain-aware verification is mentioned in HAO's blog post
\ac{hao} coordinates role-specialized agents through a facilitator (the orchestrator) that manages turn-taking, shared memory, and verification checkpoints. Agents such as \emph{PatientHistory}, \emph{Radiology}, or \emph{ClinicalTrials} focus on complementary functions aligned to \ac{mtb} workflows. The orchestrator plans and moderates the interaction, dynamically selecting only the agents needed for a case and recording an auditable trail of intermediate results. The interaction model follows a structured group-chat abstraction implemented with Semantic Kernel and Magentic-One, extended for healthcare complexity, and uses \ac{mcp} for secure, two-way tool/data connectivity.

\paragraph{Teams-native collaboration experience}
\label{subsec:frontstage}
The user experience is embedded in Microsoft Teams, where clinicians and developers converse with the orchestrator or directly \emph{@}‑mention agents inside a channel or group chat. This reduces workflow switching and lets multi-human, multi-agent conversations unfold in the same thread used for case coordination. Outputs (e.g., patient timelines or draft \ac{mtb} briefs) are distributable across Microsoft 365 apps (Word, PowerPoint) for rapid handoff to tumor board packets and follow-ups.

\paragraph{Design goals and orchestration rationale}
\label{subsec:backstage}
\ac{hao} is designed around three principles: (i) \textit{precision under specialization}, by engaging only the most relevant agents for a case; (ii) \textit{traceability}, through shared memory, inline citations, and auditable intermediate artifacts; and (iii) \textit{safety-by-design}, via domain-aware verification checkpoints and constrained tool routing. These choices aim to reduce error propagation and over-orchestration noise while preserving the benefits of division of labor across domains (history, radiology, pathology, trials). This flexibility supports diverse use cases—from rapid single-agent timelines to multi-agent workflows for complex cases—while maintaining transparency and alignment with institutional governance.

\paragraph{Real-world usage patterns} A clinical assistant can invoke the \emph{PatientHistory} agent alone to rapidly generate a concise, citation-backed timeline as an \ac{mtb} opening brief. For complex cases or deeper analysis, the orchestrator can engage multiple agents in sequence to produce a more comprehensive report. This flexibility accommodates diverse information needs: senior specialists can request clarifications during meetings, while assistants with partial context can bootstrap high-quality summaries that remain traceable to their sources.

% Early observations highlight that multi-agent orchestration introduces new complexities—such as error propagation and coordination overhead—even as it improves specialization. To mitigate these, \ac{hao} implements:
% \begin{itemize}[leftmargin=*]
%     \item \textbf{Critical checkpoints} to verify outputs from key agents before downstream use.
%     \item \textbf{Selective activation} of a few highly relevant agents per case, avoiding over-orchestration noise.
%     \item \textbf{Transparent hand-offs} that expose intermediate rationales, enabling clinicians and developers to trace how conclusions were reached and intervene when necessary.
% \end{itemize}

% \begin{figure}[t]
%     \centering
%     \includegraphics[width=\linewidth]{hao_architecture.jpg}
%     \caption{\textbf{HAO demo stack.} \emph{Left:} Teams chat with facilitator and role agents. \emph{Middle:} Orchestrator routing with selective activation and verification checkpoints. \emph{Right:} Grounding pane showing citations and lineage timeline.}
%     \label{fig:hao_stack}
% \end{figure}

\begin{figure}[t]
    \centering
    \includegraphics[width=\linewidth]{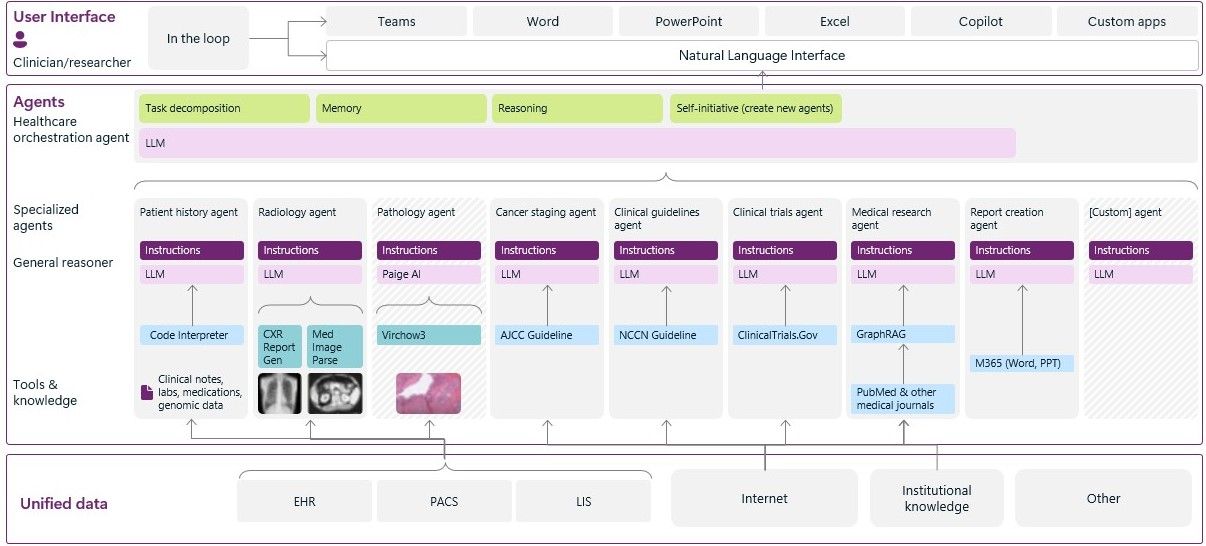}
    \caption{\ac{hao} stack diagram illustrating four layers: (i) a Teams‑native interface for clinician interaction across Microsoft 365 apps; (ii) an orchestration layer that manages task decomposition, shared memory, and verification checkpoints; and (iii) a set of role‑specialized agents paired with domain tools. These agents operate over (iv) a unified data plane integrating \ac{ehr}, PACS, LIS, institutional knowledge, and external sources.}
    \label{fig:hao_stack}
\end{figure}

\section{TBFact: Measuring Quality in \ac{mtb} Summaries}
\label{sec:tbfact}

Factuality evaluation for long, information‑dense outputs commonly adopts a \emph{decompose‑then‑verify} strategy~\cite{metropolitansky2025effectiveextractionevaluationfactual}, where choosing the right \emph{unit of analysis} is critical. Concurrent work such as FactEHR explores this paradigm for clinical notes, pairing atomic fact decomposition with \ac{llm}-based model‑as‑a‑Judge entailment to compute fact‑precision and (weighted) fact‑recall~\cite{munnangi2025factehrdatasetevaluatingfactuality}.

% In clinical text, fact decomposition is harder due to dense terminology and heterogeneous note styles~\cite{munnangi2025factehrdatasetevaluatingfactuality}
However, claim extraction introduces challenges of ensuring consistent coverage of verifiable content, maintaining entailment fidelity, and handling decontextualization~\cite{metropolitansky2025effectiveextractionevaluationfactual}. Evidence favors \emph{fine‑grained, explicitly delimited units}: on a broad benchmark, \emph{claim‑triplets} improve hallucination detection over sentence- or sub‑sentence checks with strong human alignment~\cite{hu-etal-2024-knowledge}, and in adjacent retrieval settings, \emph{proposition}‑level indexing (concise, self‑contained factoids) improves retrieval and downstream QA~\cite{chen-etal-2024-dense}. 

Motivated by these insights, TBFact adopts short, natural‑language \emph{clinical factual claims}, defined as statements that can be objectively verified as true or false based on empirical evidence or reality~\cite{ni-etal-2024-afacta}. TBFact builds on RadFact~\cite{bannur2024maira2groundedradiologyreport}, adapting its principles to text-only healthcare interactions, and introducing two key extensions: (i) importance stratification to prioritize clinically salient information, and (ii) explicit categorization of mismatches into omissions and unsupported claims. While rubric-based frameworks like HealthBench~\cite{arora2025healthbenchevaluatinglargelanguage} also aim to assess inclusion of clinically relevant aspects, TBFact achieves this through automated fact extraction rather than curated rubrics—enabling scalable, clinically meaningful evaluation without manual rubric authoring.

% TBFact builds on RadFact~\cite{bannur2024maira2groundedradiologyreport}, which introduced claim-wise factuality evaluation for radiology reports, and adapts this paradigm to text-only healthcare interactions. Our framework introduces two key extensions: (i) importance stratif

% TBFact builds on RadFact~\cite{bannur2024maira2groundedradiologyreport}, adapting its principles to text-only healthcare interactions, and introducing two key extensions: (i) importance stratification to prioritize clinically salient information, and (ii) explicit categorization of mismatches into omissions and unsupported claims. Unlike traditional lexical or semantic similarity metrics, TBFact operates over clinical factual claims, computing precision and recall via logical entailment. While rubric-based frameworks like HealthBench~\cite{arora2025healthbenchevaluatinglargelanguage} also aim to assess inclusion of clinically relevant aspects, TBFact achieves this through automated fact extraction rather than curated rubrics—enabling scalable, clinically meaningful evaluation without manual rubric authoring.

TBFact's evaluation pipeline has four main stages:
\begin{enumerate}[leftmargin=*]
    % (e.g., “EGFR L858R detected in 2023” or “Patient has BRCA1 mutation”)
    \item \textbf{Claim Extraction:} Both reference and candidate summaries are decomposed into \emph{clinical factual claims}. These units are independently verifiable and granular enough to reveal omissions or unsupported additions; no fixed ontology schema is required.
    
    \item \textbf{Importance Classification:} Each extracted fact is assigned, an importance level (\emph{high}, \emph{medium}, \emph{low}) based on \ac{mtb} salience (e.g. actionable biomarkers, staging inflections, treatment response). Importance labels are generated programmatically with an \ac{llm} rubric.

    \item \textbf{Bidirectional Entailment and Metric Aggregation:} For each extracted fact, TBFact determines whether it is \emph{fully entailed}, \emph{partially entailed}, or \emph{not entailed} by the counterpart text, then aggregates into (i) Recall, a proxy for \emph{completeness}, and (ii) Precision, a proxy for \emph{succinctness}. Partial credit (0.5) keeps the metric unsaturated and rewards incremental improvements.
    
    \item \textbf{Error Attribution:} Non‑entailed or partially entailed facts are categorized as \emph{omissions} (missed reference facts) or \emph{unsupported claims} (hallucinations/extraneous details), enabling actionable diagnostics.

    % \item \textbf{Metric Aggregation:} TBFact computes (i) Recall, a proxy for \emph{completeness}: fraction of reference facts captured by the candidate, and (ii) Precision, a proxy for \emph{succinctness}: fraction of candidate facts supported by the reference. Partial credit ensures the metric remains \emph{unsaturated}, rewarding incremental improvements rather than collapsing near-perfect outputs into a single score.
\end{enumerate}

While \acp{llm} serve as practical evaluators for complex, open-ended clinical text, their judgments are inherently subjective. Therefore, \ac{llm}-based evaluation should be treated as \emph{directional signals}, not absolute scores, and must be paired with human checks (see Section~\ref{sec:humanval}).

\section{Data and Methods}
\label{sec:data-methods}

\subsection{\emph{TB-Bench} Dataset}
For evaluation, we used \emph{TB-Bench}, a proprietary, de-identified corpus independently curated by a leading healthcare provider partner. TB-Bench aggregates longitudinal materials for 71 oncology patients discussed in \acp{mtb}, including tumor board transcripts, exported \ac{ehr} data, and clinician-authored patient summaries. Inclusion required both a transcript containing a case summary and complete EHR exports (imaging, biomarkers, diagnoses, medications, procedures, and social history).
% \footnote{Access details omitted due to institutional privacy.}

\paragraph{Reference patient summaries} Because original summaries in transcripts were short and varied in format, we applied \ac{llm}-based structuring techniques to standardize them into a template comprising a narrative paragraph followed by a chronological timeline of clinical events. This post-processing step reduces variability and ensures fair comparison.

\paragraph{Dataset-verifiable facts} During curation, discrepancies were observed between transcript summaries and the available records (e.g., references to notes or studies later excluded in the de-identification pipeline), complicating direct comparison. To avoid penalizing systems for information absent in the dataset snapshot, we restricted evaluation to \emph{dataset-verifiable facts}—claims that can be grounded in the materials included in TB-Bench. We expand on the filtering process in Appendix~\ref{app:data}. % HAO Evaluation Blog Post

\subsection{\emph{PatientHistory} agent}
The \emph{PatientHistory} agent assembles a citation-backed clinical timeline and summary as input to \ac{mtb} discussion. It builds on the \emph{Universal Abstraction} framework~\cite{wong2025universalabstraction}, a zero-shot medical abstraction approach leveraging modular prompt templates to extract oncology attributes (e.g., tumor site, histology, TNM stage, biomarkers, treatments) from unstructured clinical text without task-specific labeling. This foundation enables the agent to retrieve and structure longitudinal information while maintaining grounding to source data for auditability.

\subsection{Experiments}
We isolate the \emph{PatientHistory} agent to test whether a single specialized agent can match the quality of current \ac{mtb} summaries, which are short and variable in format. This provides a controlled baseline before exploring multi-agent workflows. 

We compare two prompting strategies: a generic baseline and a \emph{specialized} prompt that explicitly instructs chronological organization and prioritization of \ac{mtb}-critical elements (e.g., biomarkers with dates, treatment lines and responses, imaging-based disease status). Specialized prompting proved the most effective lever for improving completeness during inner-loop iterations.

\section{Results}
\label{sec:results}

\paragraph{Quantitative performance}
On \emph{TB-Bench} with specialized prompting, the \emph{PatientHistory} agent achieved \textbf{0.84 TBFact recall} on high-importance facts under strict entailment, which translates to inclusion of \textbf{94\% of high-importance information} when counting partial entailments.\footnote{“Strict” counts only fully entailed facts as correct; partial entailments receive fractional credit (0.5).} Table~\ref{tab:prompt} summarizes recall across prompting strategies. We report TBFact Recall as the primary metric because, in oncology workflows, omissions of clinically salient information are higher‑risk than inclusion of extraneous details. Complete results are reported in Appendix~\ref{app:tbfact}.
% On \emph{TB-Bench} with specialized prompting, the \emph{PatientHistory} agent captured \textbf{94\% of high-importance information} when counting partial entailments and achieved \textbf{0.84 TBFact recall} on high-importance facts under strict entailment. Table~\ref{tab:prompt} summarizes performance across prompting strategies.

\begin{table}[h]
\centering
\begin{tabular}{lcc}
\toprule
\textbf{Configuration} & \textbf{TBFact Recall (All)} & \textbf{TBFact Recall (High-importance)} \\
\midrule
Generic baseline & 0.56 & 0.66 \\
Specialized prompt & \textbf{0.71} & \textbf{0.84} \\
\bottomrule
\end{tabular}
\caption{TBFact recall of patient summaries generated by the \emph{PatientHistory} agent. Importance labels derived via \ac{llm}-based rubric.}
\label{tab:prompt}
\end{table}

Specialized prompts improved completeness by aligning outputs with reference expectations, but this also highlights a dependency on reference quality: succinct ground truths can penalize comprehensive outputs. This reinforces the need for standardized, clinically rich benchmarks. Notably, even in single-agent mode, \emph{PatientHistory} matched the information quality of current \ac{mtb} summaries, suggesting potential time savings without sacrificing completeness, and leaving room for multi-agent workflows to address more complex reasoning tasks. Appendix~\ref{app:tbfact} gives additional evaluation details.

\paragraph{Qualitative analysis}
TBFact surfaced common error modes: temporal misalignment, unsupported aggregation, and omissions tied to dataset mismatch. Partial entailments often reflected clinically relevant details absent from the reference, illustrating why partial credit is essential. Examples appear in Appendix~\ref{app:qual}.

\subsection{Human Validation Study}
\label{sec:humanval}
To establish TBFact's reliability as an evaluation metric, we conducted a comprehensive validation study with medically trained annotators across 53 double-annotated facts for human-human comparison, and 71 patient records for system-human comparison from TB-Bench. The study assessed three critical aspects of TBFact:

% TODO: Reference for atomic meaning?
\begin{itemize}[leftmargin=*]
  \item \textbf{Claim extraction validity}: Annotators verified whether system-extracted claims represented self-contained, clinically meaningful facts. We observed near-perfect inter-annotator agreement (0.999), with system extraction accuracy at 99.9\%.
  
  \item \textbf{Clinical importance classification}: Annotators categorized claims as \emph{high}, \emph{medium}, or \emph{low} importance based on clinical decision-making impact. Agreement was 0.66 (strict) and 0.77 (allowing one-level difference), with system classification matching human judgment in 60\% of cases strictly, and 93\% when allowing adjacent importance levels.
  % previously: "human judgment in 66\% of cases strictly" -- where was 66 from? it wasnt in the previous blob, was this just typo or from earlier dataset?
  
  \item \textbf{Entailment judgment}: Annotators assessed whether reference data supported each claim. Agreement reached 0.976, with system entailment decisions matching human judgment in 88\% of cases.
  %Agreement reached 0.914, with system entailment decisions matching human judgment in 88\% of cases.
  %the 0.914 was for an earlier subset - for consistency across this section we'll report the newest IAA numbers even if the sample size is smaller
\end{itemize}

Critically, end-to-end TBFact F1 scores showed strong correlations with human expert assessments: 55.8\% (Kendall), 70.5\% (Pearson), and 72.8\% (Spearman), validating TBFact as an effective proxy for expert evaluation of clinical factuality and completeness.

% \paragraph{Preliminary human validation study}
% Two trained annotators reviewed 71 patient cases to assess (a) claim extraction accuracy, (b) importance classification, and (c) entailment judgments. Agreement was near-perfect for extraction (99.9\%), substantial for importance (0.66 strict, 0.77 relaxed), and high for entailment (0.914). System accuracy for importance classification reached 93\% when allowing one-level tolerance. Correlation between TBFact F1 and human ratings was moderate-to-strong (Pearson $r=0.705$), supporting TBFact as a reliable proxy for iterative development.

\section{Discussion}
\label{sec:discussion}

The study addresses two practical pain points in molecular tumor boards (MTBs)—time burden and variable information quality—by pairing a Teams‑native agentic workflow (\ac{hao}) with a claim‑level evaluation framework (TBFact). In single‑agent mode, the \emph{PatientHistory} agent matches the information quality of current summaries, capturing 94\% of high‑importance content when partial entailments are counted and achieving 0.84 strict recall on the same subset. Because TBFact localizes errors to atomic claims and distinguishes omissions from unsupported additions, results are actionable for prompt and policy iteration. The metric’s institution‑controlled deployment further suits clinical governance: evaluation can be run locally without sharing patient text, and judgments remain auditable at the claim level. An additional practical insight is that specialized prompting improves completeness by targeting the same families of facts prioritized in reference summaries, which in turn highlights the value of standardized, clinically rich ground truths.

Findings are scoped to a single specialized agent rather than to end‑to‑end multi‑agent orchestration. TBFact relies on model‑as‑a‑judge decisions that, despite substantial agreement with human reviewers, should be treated as directional signals rather than absolute ground truth. Importance labels are \ac{llm}‑assisted and spot‑checked, not exhaustively adjudicated. Finally, evaluation is constrained to dataset‑verifiable facts within a curated, de‑identified snapshot; this avoids penalizing missing records but narrows the space of claims considered.

Several extensions are immediate: (i) per‑agent evaluation to quantify each role in isolation and, separately, end‑to‑end assessment of final \acp{mtb} recommendations against real case decisions; (ii) multi‑agent evaluation tracing cross‑agent omissions/distortions with claim‑level attribution; (iii) \ac{hao} extensions: build on existing agent-agnostic design for further support mixed human-and-agent workflows, and to automate additional clinical use cases (iv) metric refinements (weighted partial credit, category‑wise analyses for biomarkers/imaging/therapies); and (v) prospective measurement of edit burden and preparation time in live or simulated tumor boards. Overall, \ac{hao} and TBFact offers a practical baseline for MTB preparation—useful today in single‑agent mode, with clear headroom for orchestrated, domain‑specialized, multi‑modal workflows.

\bibliographystyle{unsrt}
\bibliography{main}

%%%%%%%%%%%%%%%%%%%%%%%%%%%%%%%%%%%%%%%%%%%%%%%%%%%%%%%%%%%%
\newpage

\appendix

\section*{Appendix}
% \addcontentsline{toc}{section}{Appendix}
\section{Importance-stratified TBFact Results}
\label{app:tbfact}

Table~\ref{tab:tbfact_results} gives the full TBFact evaluation results on the \emph{PatientHistory} agent summary using the ``specialize'' tumor board prompts. As discussed in Section ~\ref{sec:results}, we consider TBFact recall to be the most important metric, ensuring that facts deemed important by a human were also included by the agent. By stratifying based on fact importance, we can see that the agent and humans have higher levels of agreement on the most critical facts to include in the summary.

The TBFact precision metric can be seen as a measure of ``succinctness'', ensuring that the agent does not include large amounts of irrelevant information. Both the precision and number of facts (\emph{p\_support} and \emph{r\_support}) show that the agent produces more overall facts than the human, indicating that there are additional improvements that can be made to align the agent and human understanding of facts important to the \ac{mtb} process.

\begin{table}[ht]
\centering
\begin{tabular}{lcccccc}
\hline
Importance & Precision & Recall & F1 & p\_support & r\_support \\
\hline
Overall    & 0.446     & 0.711  & 0.548 & 4322 & 1573 \\
High       & 0.616     & 0.838  & 0.710 & 2219 & 816 \\
Medium     & 0.289     & 0.644  & 0.399 & 1550 & 517 \\
Low        & 0.200     & 0.422  & 0.272 & 553  & 240 \\
\hline
\end{tabular}
\caption{Overall and importance-stratified TBFact results for the specialized prompts. \emph{p\_support} and \emph{r\_support} give the number of facts identified in the agent- and human-generated summaries, respectively.}
\label{tab:tbfact_results}
\end{table}

\section{Dataset-Verifiable Fact Filtering}
\label{app:data}

To ensure fair evaluation under conditions where complete patient records are available, we restrict the reference set to \emph{dataset-verifiable facts}—claims from the reference summary that can be grounded in the TB-Bench dataset (EHR exports and clinical notes). This prevents penalizing systems for omissions caused by de-identification or missing records.

Each reference fact is checked for support in the dataset using an \ac{llm}-based entailment check (GPT-4.1, temperature 0.0). Facts labeled \texttt{Yes} or \texttt{Partial} are retained, and unsupported facts are excluded before metric computation. This step ensures TBFact metrics reflect only claims verifiable within the dataset snapshot. Table~\ref{tab:filtering} shows illustrative exclusions.

\algrenewcommand\algorithmicrequire{\textbf{Input:}}
\algrenewcommand\algorithmicensure{\textbf{Output:}}

\begin{algorithm}[h]
\caption{Dataset-Verifiable Fact Filtering}
\label{alg:filtering}
\begin{algorithmic}[1]
\Require Patients $P$; gold facts per patient $G[p]$; curated notes per patient $N[p]$
\Ensure Filtered gold facts per patient $F[p]$
\For{each patient $p \in P$}
  \State $notes \gets N[p]$ \Comment{All \emph{TB-Bench} notes for patient $p$}
  \State $facts \gets G[p]$ \Comment{Facts extracted from reference summary for patient $p$}
  \State $supportMask \gets$ array of \texttt{False} of length $|facts|$
  \For{each $note \in notes$}
    \State $entailments \gets \Call{LLMJudge}{facts, note}$ \Comment{labels: Yes/No/Partial}
    \For{$i \gets 1$ \textbf{to} $|facts|$}
      \If{$entailments[i] \in \{\texttt{Yes}, \texttt{Partial}\}$}
        \State $supportMask[i] \gets \texttt{True}$
      \EndIf
    \EndFor
  \EndFor
  \State $F[p] \gets \{\, facts[i] \mid supportMask[i] = \texttt{True} \,\}$
\EndFor
\State \Return $F$
\label{algo:datafilter}
\end{algorithmic}
\end{algorithm}

\begin{table}[h]
\centering
\begin{tabular}{p{0.4\linewidth}p{0.5\linewidth}}
\toprule
\textbf{Excluded Fact} & \textbf{Reason for Exclusion} \\
\midrule
``Patient reported fatigue during CCNU cycle'' & Symptom mentioned in transcript but absent from EHR export \\
``Plan to continue CCNU and monitor with imaging'' & No corresponding plan documented in available records \\
\bottomrule
\end{tabular}
\caption{Illustrative examples of excluded facts.}
\label{tab:filtering}
\end{table}

Using Algorithm~\ref{alg:filtering}, we find that 75.1\% of facts from the tumor board transcripts are supported by the available patient notes, and are retained in the final evaluation.

\subsection{Hallucination Detection}

The TBFact precision metric penalizes facts in the agent-generated summary that are not verifiable given the human-generated summary. These facts can be extraneous information that exists in the patient's history but which a human decided not to include. However, it is also possible that these are hallucinations that are not supported by available clinical notes. In order to identify possible hallucinations, we apply Algorithm~\ref{alg:filtering} to the agent-generated summary, evaluating whether each fact is supported by the patient's clinical notes. We find that 96.3\% of the facts are supported by the patient notes. Manual inspection of the remaining 3.7\% of the facts shows that the majority are cross-note inferences such as ``the patient has undergone multiple resections'' that are verifiable from the complete history, but not supported by a single clinical note.

\section{Qualitative Examples of TBFact Evaluation}
\label{app:qual}
Table~\ref{tab:qualitative} illustrates how TBFact evaluates factual alignment between agent-generated summaries and reference data, including entailment judgments and explanations.

\begin{longtable}{@{}p{0.33\textwidth} p{0.37\textwidth} p{0.26\textwidth}@{}}
\toprule
\textbf{Facts Extracted from Agent Response} & \textbf{Related Excerpt from Reference Text (Ground Truth)} & \textbf{TBFact Judgment} \\
\midrule
Molecular studies from the 2019-05-18 surgery identified TERT promoter mutation, PTEN mutation, EGFR amplification, CDKN2A/B deletion, monosomy 10, and trisomy 7. &

[…] \par\smallskip \textbf{Tumor Genetics:} \par - EGFR: Amplified \par - CDKN2A/B: Deleted \par - PTEN: p.L112R \par - TERT: c.-146C>T \par - Chromosome 10: Monosomy \par - Chromosome 7: Trisomy \par\smallskip […] \par\smallskip \textbf{Timeline:} \par\smallskip - 05/18/2019: Diagnosis of multifocal glioblastoma; craniotomy and resection of lesion from right temporal lobe. \par\smallskip […] &

\textbf{Entailed:} \par\smallskip The summary lists TERT mutation, PTEN mutation, EGFR amplification, CDKN2A/B deletion, monosomy 10, and trisomy 7. \\
\midrule
Immunohistochemistry from 2019-05-18 showed GFAP positive, BRAF V600E negative, IDH1 R132H negative, ATRX retained, p53 negative, and a Ki-67 index of 3\%. &

[…] \par\smallskip \textbf{Tumor Genetics:} \par\smallskip - IDH1: Wildtype \par - BRAF V600E: Negative \par\smallskip […] \par\smallskip \textbf{Timeline:} \par\smallskip - 05/18/2019: Diagnosis of multifocal glioblastoma; craniotomy and resection of lesion from right temporal lobe. \par\smallskip […] &

\textbf{Partial Entailment:} \par\smallskip Some IHC findings match (BRAF negative, IDH1 wildtype) but others (GFAP, p53, Ki-67) are not mentioned in the reference summary. \\
\midrule
During the first cycle of CCNU on 2020-04-14, the patient reported significant fatigue, thrombocytopenia, and occasional confusion. &

\textbf{Introduction:} \par\smallskip […] The patient is experiencing poor tolerance to lomustine and is considering discontinuation due to further disease progression as confirmed by recent MRI scans. […] \par\smallskip \textbf{Timeline:} \par\smallskip - 04/14/2020 - Present: Lomustine treatment initiated. \par\smallskip […] &

\textbf{Partial Entailment:} \par\smallskip Poor tolerance to lomustine is reported, but specific side effects are not listed in the reference summary. \\
\midrule
On 2020-05-16, the plan was to continue CCNU and monitor with imaging. &

No related information in the reference text. &

\textbf{No Entailment:} \par\smallskip No mention in the summary of a plan on 2020-05-16 to continue CCNU with imaging follow-up. \\
\bottomrule
\caption{Examples of facts extracted from the \emph{PatientHistory} response, along with relevant excerpt from the reference patient summary and \emph{TBFact}'s entailment analysis result for the respective fact.}
\label{tab:qualitative}
\end{longtable}

\section{Contributors}
\label{app:contributors}

This work reflects a broad, coordinated effort across engineering, research, and clinical partners. We gratefully acknowledge the many individuals whose expertise, dedication, and collaboration were essential to building, refining, and delivering \ac{hao} and its evaluation framework. The following contributors are listed alphabetically.

\medskip
\noindent\textbf{Contributors}

\begin{multicols}{2}
\begin{itemize}[leftmargin=*,label={},nosep]
  \item Aiden Gu
  \item Alberto Santamaria-Pang
  \item Alexander Ersoy
  \item Alyssa Unell
  \item Arun C K
  \item Chris Burt
  \item Evan Miller
  \item Ivan Tarapov
  \item Jameson Merkow
  \item Kumar Thirumalaiah
  \item Manoj Kumar
  \item Mu Wei
  \item Naiteek Sangani
  \item Naveen Valluri
  \item Saumil Shrivastava
  \item Vincent Fitzgerald
  \item Will Guyman
\end{itemize}
\end{multicols}

%%%%%%%%%%%%%%%%%%%%%%%%%%%%%%%%%%%%%%%%%%%%%%%%%%%%%%%%%%%%

\end{document}